\newcommand{\CLASSINPUTtoptextmargin}{0.75in}
\newcommand{\CLASSINPUTbottomtextmargin}{0.75in}
\newcommand{\CLASSINPUTinnersidemargin}{0.75in}
\newcommand{\CLASSINPUToutersidemargin}{0.75in}
\documentclass[letterpaper,10pt,conference]{IEEEtran}
\IEEEoverridecommandlockouts

\usepackage[T1]{fontenc}
\usepackage{newtxtext,newtxmath}
\usepackage{amsmath}
\usepackage{graphicx}
\usepackage{booktabs,tabularx,array}
\usepackage{cite}
\usepackage{microtype}
\usepackage{algorithm}
\usepackage{algpseudocode}
\usepackage{stfloats}
\usepackage{placeins}
\usepackage{balance}
\usepackage{etoolbox}
\usepackage{needspace}

\apptocmd{\thebibliography}{%
  \interlinepenalty=0\relax
  \clubpenalty=10000\relax
  \widowpenalty=10000\relax
}{}{\PackageError{CommitFlow-layout}{Bibliography layout patch failed}{Check the bibliography environment.}}

\renewcommand{\IEEEtitletopspaceextra}{0.29in}

\usepackage[hidelinks]{hyperref}
\hypersetup{pdftitle={CommitFlow: Semantic Commitment Verification and Local Correction for Long-Horizon Robot Manipulation VLA Execution},pdfauthor={Zixiang Zhao, Yansong Feng, Jibao Yuan, Yang Yang, Chaoyu Wang, Haoran Xiao, Jianjun Ma, Hui Zhang, Chuang Cheng}}
\makeatletter
\let\CF@makecaption\@makecaption
\long\def\@makecaption#1#2{%
  \ifx\@captype\@IEEEtablestring
    \CF@makecaption{#1}{#2}%
  \else
    \begingroup\let\footnotesize\normalsize
    \CF@makecaption{#1}{#2}\endgroup
  \fi}
\newcommand{\fs@CFruled}{\fs@ruled
  \def\@fs@pre{\kern6pt\hrule height.8pt depth0pt\kern2pt}}
\makeatother
\floatstyle{CFruled}
\restylefloat{algorithm}

\newcolumntype{Y}{>{\raggedright\arraybackslash}X}
\newcommand{\thd}[1]{\shortstack{#1}}

\newcommand{\Keep}{\textsc{Keep}}
\newcommand{\Correct}{\textsc{Correct}}
\newcommand{\Hold}{\textsc{Hold}}
\newcommand{\R}{\mathbb{R}}
\newcommand{\ind}{\mathbb{I}}
\newcommand{\relu}[1]{\left[#1\right]_{+}}

\DeclareMathOperator{\vecop}{vec}
\title{\fontsize{16}{20}\selectfont\bfseries CommitFlow: Semantic Commitment Verification and Local Correction for Long-Horizon Robot Manipulation VLA Execution}
\author{%
\IEEEauthorblockN{Zixiang Zhao, Yansong Feng, Jibao Yuan, Yang Yang, Chaoyu Wang,\\
Haoran Xiao, Jianjun Ma, Hui Zhang, and Chuang Cheng\textsuperscript{*}}%
\IEEEauthorblockA{National University of Defense Technology, Changsha, China}%
\thanks{\textsuperscript{*}Corresponding author: Chuang Cheng.}%
}

\makeatletter
\newcommand{\CFteaser}{%
  \vspace{-12pt}
  \begin{minipage}{\textwidth}
    \centering
    \includegraphics[width=0.96\textwidth]{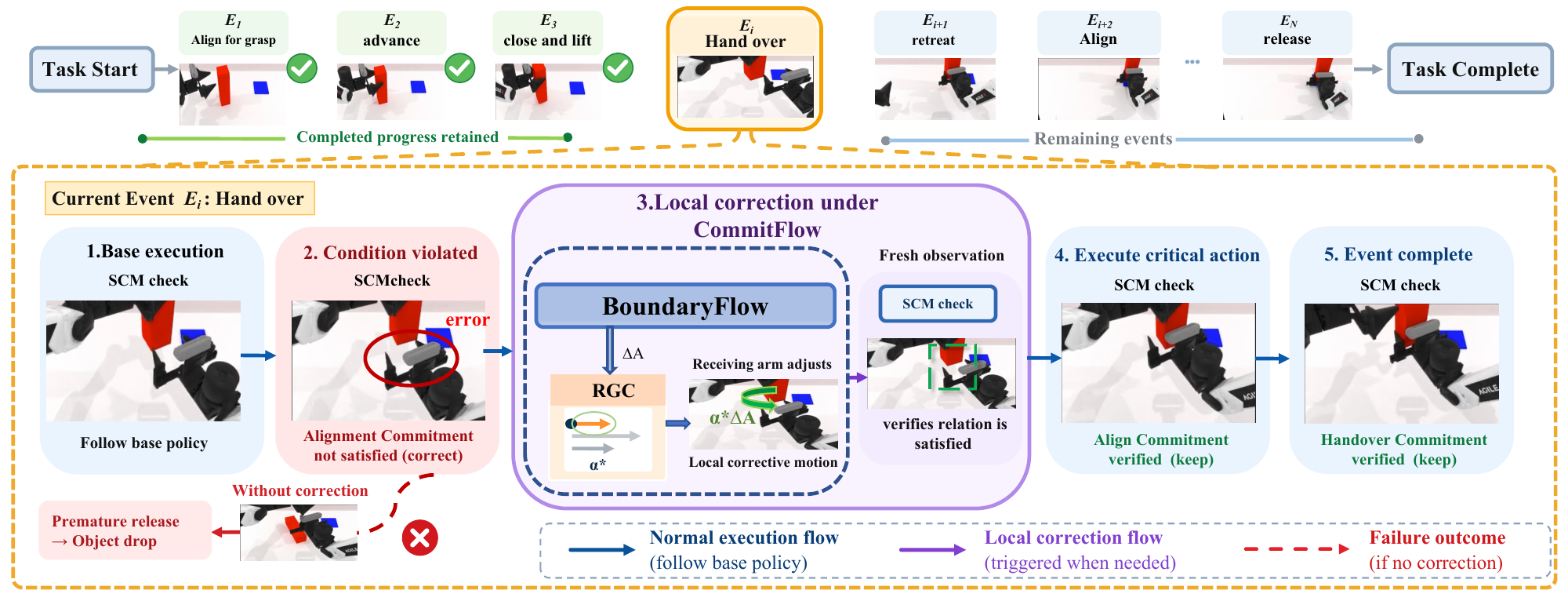}
    \def\@captype{figure}
    \caption{\textbf{Event progression and local correction in CommitFlow.}
    During handover, the sender retains the object while the receiver adjusts its alignment. BoundaryFlow supplies a local correction, RGC scales it, and SCM re-verifies the physical conditions from fresh observations before dependent actions resume.}
    \label{fig:lifecycle}
  \end{minipage}\par\vspace{0.7\baselineskip}}
\makeatother
\IEEEaftertitletext{\CFteaser}

\begin{document}
\maketitle
\thispagestyle{empty}
\pagestyle{empty}

\begin{abstract}
Although vision-language-action (VLA) policies have advanced rapidly, long-horizon execution may still progress to the next task stage before the required physical effect has been established. We call this a mismatch between semantic commitments---physical conditions that a stage must establish or maintain---and the actual physical state. Because an action command alone cannot confirm such a condition, local deviations can propagate and cause task failure. To address this problem, we present CommitFlow, a closed-loop execution framework that combines commitment monitoring with local correction while keeping the base policy frozen. CommitFlow integrates three components. A Semantic Commitment Monitor (SCM) compares stage requirements against current state evidence and holds back dependent actions when a required condition is unmet or violated. BoundaryFlow then generates a local correction conditioned on the current state and base action, and Relation and Gain Calibration (RGC) selects the smallest correction strength that satisfies the relevant constraints. Across the ten common RoboTwin 2.0 benchmark tasks, CommitFlow achieves a mean success rate of 75.9\%, improving on the base policy $\pi_{0.5}$ by 22.7\%. Cross-policy experiments show consistent gains, pointing toward reliable long-horizon robot execution.
\end{abstract}

\section{Introduction}
Vision-Language-Action (VLA) models~\cite{zitkovich2023rt2,kim2025openvla} map visual observations, language instructions, and robot states directly to robot actions, enabling robots to perform a wide range of manipulation tasks. However, extending VLA models to long-horizon tasks remains challenging. Unlike short-horizon tasks, long-horizon tasks typically involve multiple interdependent execution stages, where the outcome of the current stage determines whether subsequent stages can proceed correctly. More specifically, during execution, the current stage may fail to achieve its intended physical effect, while the policy continues to advance through the task sequence and may even transition to the next stage before the current one is properly completed. In such cases, subsequent actions are executed from an incomplete or incorrect physical state, causing local execution deviations to accumulate and propagate across stages, ultimately leading to task failure.

This mismatch can arise when stage transitions occur before physical preconditions are met or when concurrent actions interfere. Figure~\ref{fig:lifecycle} illustrates the handover case: the sender retains the object while the receiver corrects its alignment, and release waits for verified receiver grasp. Other failures include release outside the placement region or premature basket grasping during ongoing placement. Reliable execution therefore requires verification of physical preconditions and action dependencies before advancing.

To address this issue, existing studies typically detect anomalies using policy features, visual evidence, or constraint monitoring, and recover execution through replanning or retries. However, these methods often intervene only after clear signs of task failure have emerged. At that point, recovery must deal with a more complex failure state and may also disrupt previously established grasping, support, or interaction relationships. Therefore, we argue that reliable long-horizon execution should not wait until local deviations have evolved into task failures before initiating recovery. Instead, intervention should be moved ahead of critical stage transitions, so that deviations can be identified and corrected before adverse consequences arise and propagate to subsequent stages.

CommitFlow is built around this principle. It represents the physical conditions required for stage progression as semantic commitments and organizes execution as a closed loop of verification, correction, and re-verification. The Semantic Commitment Monitor (SCM) evaluates whether the stage-specific physical conditions are satisfied based on the current observation. When a required condition is not met, SCM postpones downstream actions that depend on it while preserving unaffected execution progress. For the detected local deviation, BoundaryFlow then corrects the action proposed by the base policy without replacing the underlying task policy. The Relation and Gain Calibration (RGC) module further determines the necessary correction magnitude according to the current relational deviation, thereby avoiding excessive intervention. After the correction is executed, SCM re-evaluates the relevant conditions using the updated observation, and task execution proceeds only when the corresponding semantic commitment has been re-established.

Overall, this work addresses early intervention and local correction for long-horizon VLA execution through three aspects:

\noindent 1.\ \textbf{Semantic commitment monitoring.} We propose SCM to identify unsatisfied physical conditions through state estimation and commitment verification, defer dependent actions, and preserve valid progress.

\noindent 2.\ \textbf{Learned local correction and gain calibration.} We propose BoundaryFlow to learn local updates from base actions and expert corrections, and RGC to build stage-wise reference ranges from successful data and select correction strength.

\noindent 3.\ \textbf{Cross-policy evaluation and correction reuse.} We evaluate performance across different base policies and examine the transfer of correction strategies to similar tasks.

\Needspace{6\baselineskip}
\section{Related Work}
Research on reliable long-horizon VLA execution has mainly progressed along three directions: monitoring task states and physical conditions to assess execution status, handling execution deviations through active recovery and learned action correction, and directly guiding or adjusting actions at inference time. Together, these studies extend the ability of VLA policies to perceive and intervene during execution from different perspectives.

\subsection{Task-State Perception and Physical-Condition Monitoring}
SAFE estimates cross-task failure risk from internal VLA representations~\cite{safe}. DoReMi formulates plan-execution inconsistencies as constraint violations~\cite{doremi}, while ReViP addresses false completion before the intended goal has been achieved~\cite{revip}. More recently, PhysReflect-VLA evaluates candidate-action feasibility during execution and uses structured self-reflection to guide subsequent action sampling~\cite{physreflectvla}. These studies advance execution monitoring from global risk estimation toward explicit execution-state verification. CommitFlow focuses on stage-dependent physical requirements represented as commitments that gate dependent actions and are re-verified after localized correction.

\subsection{Active Recovery and Learned Action Correction}
Once an execution deviation has been detected, restoring task progress constitutes another major line of research.

REFLECT uses multimodal execution history to interpret failures and guide subsequent planning~\cite{reflect}, while AIC adjusts manipulation poses online through interactive feedback~\cite{aic}. FPC-VLA introduces a VLM-based supervisor that evaluates action viability at keyframes and generates language corrections specifying direction and magnitude when potential failures are identified~\cite{fpcvla}. For longer-horizon tasks, CycleVLA combines progress estimation with termination prediction and uses backtracking near stage transitions~\cite{ma2026cyclevla}, while FLARE handles deviations through Retry and Reset behaviors~\cite{flare}. RePO-VLA further learns corrective behaviors from successful, failed, and recovery trajectories~\cite{repovla}. Subtask-level backtracking or skill resets can also roll back execution components that remain valid, potentially disrupting established progress. FPC-VLA anticipates action failure at keyframes; CommitFlow instead verifies the current stage's physical commitments and selectively withholds dependent actions until the required conditions are established.

\begin{figure*}[!t]
\centering
\includegraphics[width=0.90\textwidth]{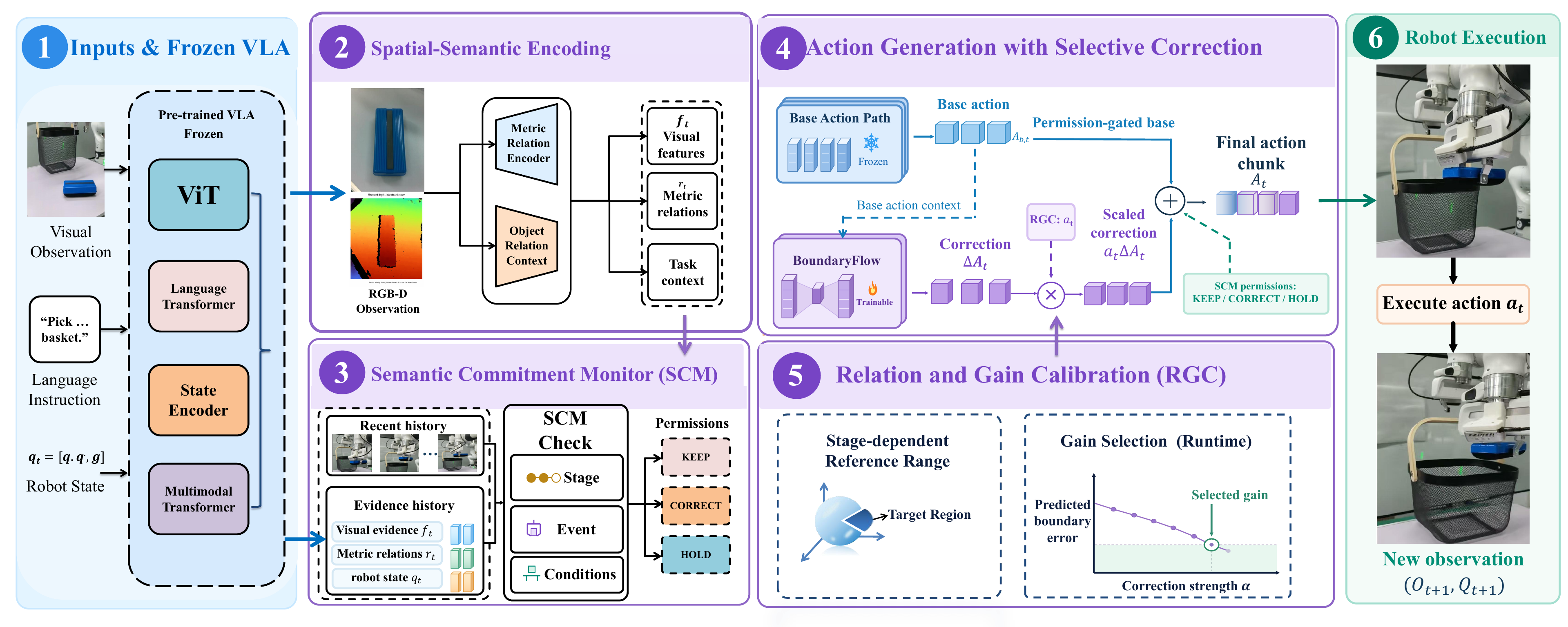}
\caption{\textbf{CommitFlow overview.} RGB-D encoding supplies visual features, metric relations, and task context. SCM assigns action permissions; BoundaryFlow predicts a base-action-conditioned residual; RGC selects its strength. The permission-gated action is executed, and fresh observations support re-verification.}
\label{fig:overview}
\end{figure*}

\subsection{Inference-Time Action Guidance and Execution Adjustment}
Beyond explicit monitoring and recovery, another line of work directly improves actions generated by the base policy at inference time. V-GPS ranks candidate actions using an offline value function~\cite{vgps}, TACO selects more reliable action chunks according to support from successful demonstrations~\cite{taco}, and CAPS corrects drifting trajectories through local sampling and adaptive search~\cite{caps}. VLA-Corrector detects persistent mismatch between predicted and observed visual evolution, truncates stale actions, and triggers corrective replanning with an adaptive action horizon~\cite{vlacorrector}. Its VLA backbone remains frozen while an external corrector guides the recovery query. These methods improve execution through action selection, local search, or event-triggered replanning. However, trajectory drift and action quality do not by themselves establish whether stage-transition prerequisites are satisfied or which dependent actions should remain withheld until verification.

\section{Method}
Our goal is to enable a VLA policy to identify and correct deviations between intended effects and the actual state during execution, prevent task advancement before the necessary physical conditions are satisfied, and thereby improve system generalization and robustness. To this end, we propose CommitFlow, a closed-loop execution framework built around \emph{semantic commitments} that converts physical-condition deviations detected at runtime into correction targets constrained by current task requirements. SCM verifies the active commitment through lightweight temporal state estimation and passes the missing conditions and existing conditions that must be preserved to BoundaryFlow, enabling it to adjust only the affected portions of the base action. Subsequently, RGC evaluates the predicted states corresponding to different correction strengths using stage reference ranges established from successful execution data and selects a candidate that satisfies both local geometric requirements and action constraints. After the correction is executed, SCM re-verifies the relevant conditions based on new observations and accordingly determines whether to continue correction or resume subsequent actions. For different scenes with the same task structure, the task adapter binds objects and manipulation roles and updates relative states, enabling the commitment-verification and local-correction process to be reused across scene configurations.

\begin{figure*}[!t]
\centering
\includegraphics[width=0.90\textwidth]{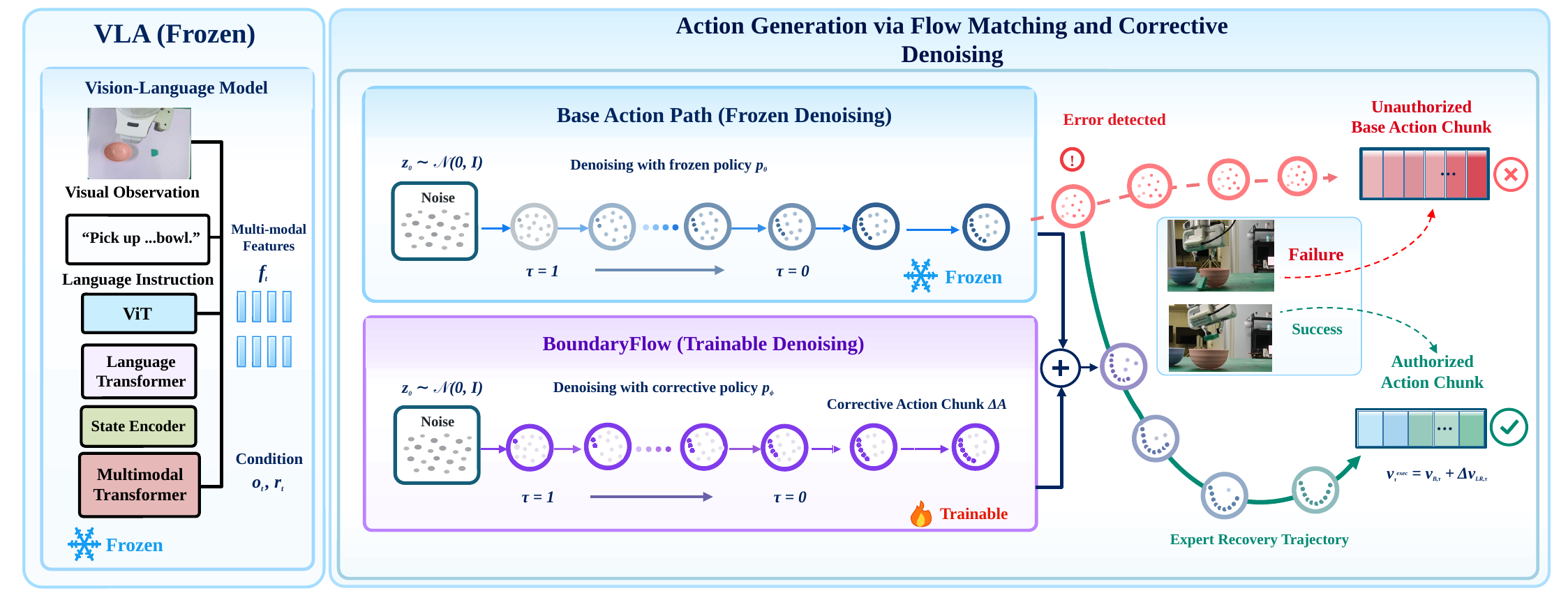}
\caption{\textbf{BoundaryFlow action generation and local correction.} The frozen VLA generates a base action chunk through flow matching, while the trainable BoundaryFlow branch generates a corrective action chunk. Combining the two produces the corrected action chunk.}
\label{fig:correction}
\end{figure*}

Figure~\ref{fig:overview} summarizes the overall architecture, and Fig.~\ref{fig:correction} illustrates how BoundaryFlow combines a local correction with the frozen policy's base action.

A frozen policy receives observations $o_t$, measured robot state $q_t$, and instruction $\ell$, and proposes
\begin{equation}
 A_{B,t}=\pi_B(o_t,q_t,\ell)\in\R^{H\times D_a}.
 \label{eq:base}
\end{equation}
Here $H$ is the horizon and $D_a$ is the action dimension. A \emph{base candidate} is this not-yet-fully-executed chunk. The runtime dispatches only an approved prefix before re-observation.

\subsection{Task-Relevant State and Event Representation}
\label{sec:perception}
Task-relevant RGB-D features $f_t$, measured joints, and kinematics provide state evidence. An adapter defines templates for pre-alignment, reaching, grasp closure, and lifting, specifying objects, arm roles, physical conditions, geometric relations, and subsequent actions. After binding perceived objects to the active template, it computes
\begin{equation}
 r_t^i=(R_i^w)^\top(p_t^{\mathrm{tgt}}-p_t^{\mathrm{src}})\in\R^3,
 \label{eq:relation}
\end{equation}
where $R_i^w$ is the event-frame orientation and source/target are interaction reference points estimated from RGB-D and kinematics. Orientation and other geometry are represented separately. Visual features condition SCM and BoundaryFlow; metric relations support RGC checks and calibration in physical coordinates. Figure~\ref{fig:scm-monitoring}(a) reports RGB-D/CAD position errors over all closed-loop frames and the ALIGN TO GRASP subset.

\subsection{Semantic Commitment Monitoring (SCM)}
\label{sec:runtime}
SCM combines a causal state estimator with a commitment verifier. Stage and event predictions establish context; only verified physical conditions authorize progression.

\textbf{Causal state estimation.}
A gated recurrent unit (GRU) encodes an $L$-frame history:
\begin{equation}
 x_t=[f_t;\,r_t^i;\,q_t],\qquad
 h_t=\operatorname{GRU}(x_{t-L+1:t}).
 \label{eq:temporal}
\end{equation}
Stage and event softmax heads read out $h_t$, while a sigmoid head predicts holding, support, contact, and release probabilities. Event predictions are matched to the adapter's active roles and legal task structure.

\textbf{History-window supervision.}
Annotated causal windows train the observer with $\mathcal L_{\mathrm{obs}}=\mathcal L_{\mathrm{stage}}+\beta_e\mathcal L_{\mathrm{event}}+\beta_p\mathcal L_{\mathrm{condition}}$: stage/event cross-entropy and physical-condition binary cross-entropy (BCE). Rule-generated labels supply supervision only. Invalid annotations are masked; condition BCE averages valid labels and is zero when none are available. Observer training is separate from \eqref{eq:loss}.

\textbf{Commitments and permissions.}
For event $e_t$, SCM checks the adapter's required conditions $\mathcal C_{e_t}$ against state estimates, metric relations, and history. Conditions are \emph{pending} until satisfied, \emph{blocking} when required by a dependent action but unsupported by evidence, and \emph{violated} after losing satisfaction. Stage predictions cannot override unresolved commitments.

Before protected actions, unsupported critical actions and their dependent time--channel elements form $\mathcal H_t$; the locally adjustable prefix forms $\mathcal C_t$:
\begin{equation}
 \Gamma_{t,k,d}=\begin{cases}
 \Hold,&(k,d)\in\mathcal H_t,\\
 \Correct,&(k,d)\in\mathcal C_t\setminus\mathcal H_t,\\
 \Keep,&\text{otherwise}.
 \end{cases}
 \label{eq:permissions}
\end{equation}
\Hold\ takes priority over local correction, while \Keep\ retains unaffected base motion.

\begin{figure}[!ht]
\centering
\includegraphics[width=\columnwidth]{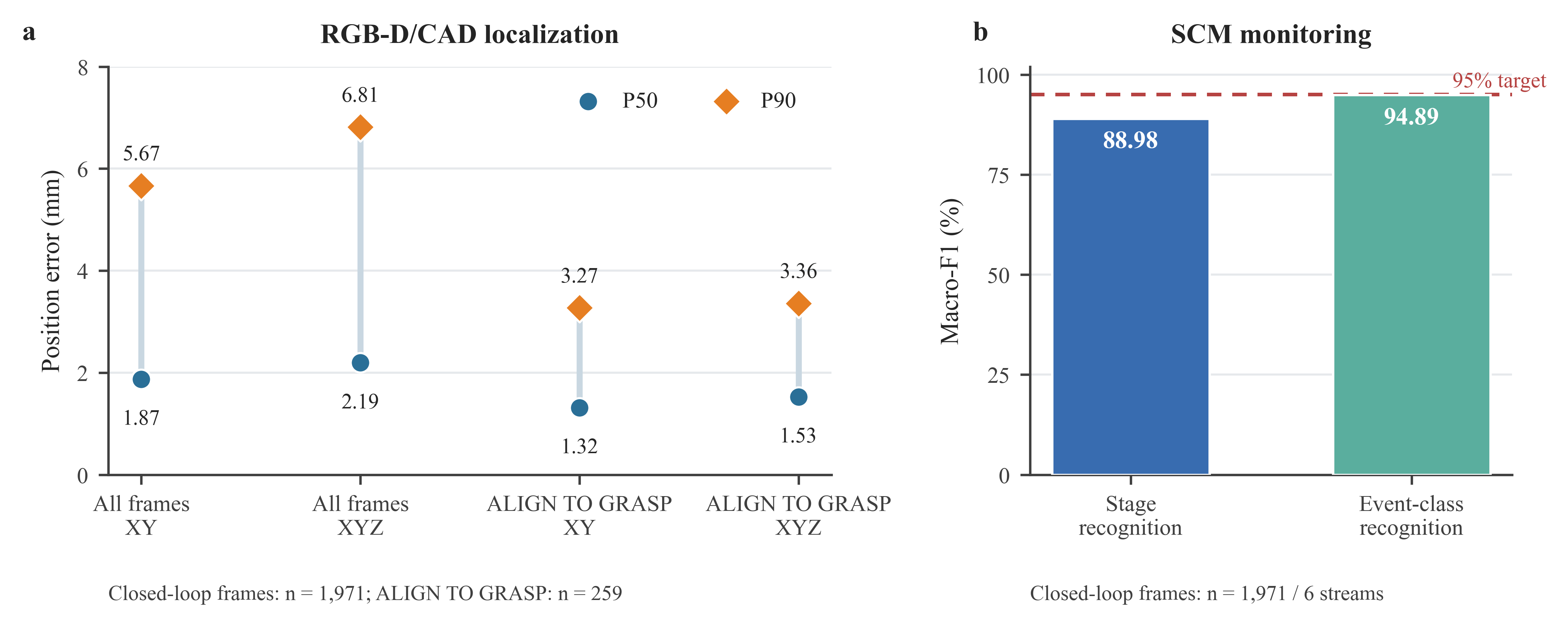}
\caption{\textbf{RGB-D/CAD localization and SCM recognition performance.} (a) Median (P50) and 90th-percentile (P90) position errors in XY and XYZ over all 1,971 closed-loop frames and the 259-frame ALIGN TO GRASP subset. (b) Macro-F1 for stage and event-class recognition across six streams (1,971 frames). The dashed line marks the 95\% target.}
\label{fig:scm-monitoring}
\end{figure}

\textbf{SCM monitoring.} Across 1,971 closed-loop frames from six streams, SCM achieves Macro-F1 scores of \textbf{88.98\%} for stage recognition and \textbf{94.89\%} for event-class recognition (Fig.~\ref{fig:scm-monitoring}(b)).

\subsection{BoundaryFlow: Base-Action-Conditioned Correction}
\label{sec:correction}
BoundaryFlow predicts bounded local residuals conditioned on the current state and frozen base proposal.

\textbf{Conditioning and low-rank correction.}
The condition $z_t$ combines local visual features, normalized object--arm relations and their protected-boundary predictions, measured joints, the base prefix, and stage/role evidence. Conditioning on the proposed prefix accounts for its effect on the deviation. A fixed active-arm mapping provides shared six-joint coordinates (superscript $c$) over $K=10$ steps for compatible arm roles.

An uncentered singular value decomposition of vectorized training expert--base residuals provides the leading $r=12$ left singular vectors $B_r\in\R^{60\times r}$. The residual is
\begin{equation}
 \begin{aligned}
 c_t&=c_{\max}\tanh f_\theta(z_t),\\
 \Delta A_t^c&=\operatorname{reshape}(B_rc_t,K,6).
 \end{aligned}
 \label{eq:head}
\end{equation}
The orthonormal basis restricts residual directions to the training subspace and yields $\|\vecop(\Delta A_t^c)\|_2\leq\sqrt r\,c_{\max}$ in normalized action coordinates.

\textbf{Boundary-aligned residual supervision.}
For each recoverable pre-transition perturbation, the frozen base and expert generate paired prefixes from the same state, $A_{B,t}^c$ and $A_t^{\star,c}$. The expert repairs the relation while preserving invariants $I_i$ and withholding the protected effect. With $M^{\mathrm{corr}}=\ind[\Gamma=\Correct]$, the coefficient target is
\begin{equation}
 c_t^\star=B_r^\top\vecop\!\left(
 M_t^{\mathrm{corr},c}\odot[A_t^{\star,c}-A_{B,t}^c]\right).
 \label{eq:expert-coefficients}
\end{equation}
Valid nominal states at the same stage anchors receive zero-correction supervision.

\textbf{Joint action and geometric training.}
We supervise coefficients against $c_t^\star$ with $\mathcal L_{\mathrm{coef}}$, corrected actions against the expert on permitted coordinates with $\mathcal L_{\mathrm{act}}$, and denoising at the flow interface with $\mathcal L_{\mathrm{denoise}}$. For geometric supervision, FK in physical joint coordinates yields $\widehat r_{t,k}^{\mathrm{corr}}$. Using the RGC margin from \eqref{eq:margin}, let $m_{t,k}^{\mathrm{corr}}=m_i(\widehat r_{t,k}^{\mathrm{corr}})$, $m_{t,0}=m_i(r_t^i)$, and $\overline m_{t,k}=(1-k/K)m_{t,0}+(k/K)\delta$. We penalize terminal error and progress below this soft reference:
\begin{equation}
 \begin{aligned}
 \mathcal L_{\mathrm{contract}}
 &=\relu{\delta-m_{t,K}^{\mathrm{corr}}}\\
 &\quad+\frac{\lambda_p}{K}\sum_{k=1}^{K}
 \relu{\overline m_{t,k}-m_{t,k}^{\mathrm{corr}}}.
 \end{aligned}
 \label{eq:contract-loss}
\end{equation}
Each term becomes zero when its margin requirement is met.

Nominal pairs use $\mathcal L_{\mathrm{null}}=\|c_t\|_2^2$. The full objective is
\begin{equation}
 \begin{aligned}
 \mathcal L_{\mathrm{BoundaryFlow}}
 &=\mathcal L_{\mathrm{coef}}+\lambda_a\mathcal L_{\mathrm{act}}
   +\lambda_d\mathcal L_{\mathrm{denoise}}\\
 &\quad+\lambda_g\mathcal L_{\mathrm{contract}}
   +\lambda_0\mathcal L_{\mathrm{null}}.
 \end{aligned}
 \label{eq:loss}
\end{equation}
Repair and null supervision use their respective sample subsets.

\textbf{Integration with base action generation.}
For $M^{\mathrm{auth}}=\ind[\Gamma\neq\Hold]$ and candidate gain $\alpha$, SCM permissions constrain composition:
\begin{equation}
 \begin{aligned}
 A_t^{\mathrm{gate}}&=M^{\mathrm{auth}}\odot A_{B,t}
 +(1-M^{\mathrm{auth}})\odot A_t^{\mathrm{hold}},\\
 A_t^{(\alpha)}&=A_t^{\mathrm{gate}}
 +\alpha M^{\mathrm{corr}}\odot I_a(\Delta A_t^c).
 \end{aligned}
 \label{eq:composition}
\end{equation}
The map $I_a$ restores full action coordinates. $A_t^{\mathrm{hold}}$ retains the required gripper state and measured joint positions (or zero increments/velocities); RGC selects $\alpha$ via \eqref{eq:feasible}.

For flow policies, apply \eqref{eq:composition} to the clean-action estimate $\bar A_B^\tau=x_\tau-\tau v_B^\tau$, with noisy action $x_\tau$, then map $\bar A_{\mathrm{exec}}^\tau$ back to the sampling field:
\begin{equation}
 v_{\mathrm{exec}}^\tau
 =\frac{x_\tau-\bar A_{\mathrm{exec}}^\tau}{\tau}
 =v_B^\tau-\frac{\bar A_{\mathrm{exec}}^\tau-\bar A_B^\tau}{\tau},
 \quad\tau>0.
 \label{eq:flow}
\end{equation}
For $\tau>0$, the mapping recovers $v_B^\tau$ whenever $\bar A_{\mathrm{exec}}^\tau=\bar A_B^\tau$.

SCM re-verifies stage conditions and retained states from fresh measurements before discarding stale actions and re-querying the base policy (Algorithm~\ref{alg:runtime}).

\subsection{Relation and Gain Calibration (RGC)}
\label{sec:regions}
\label{sec:verification}
RGC calibrates reference regions offline and selects gain along BoundaryFlow's predicted residual.

\textbf{Successful-state reference regions.}
Successful boundary states are grouped by approach direction and action configuration, retaining grasp-specific labels and normal pre-approach axial stand-off. A fitting split provides coordinatewise medians $\mu_i$ and per-axis standard deviations $S_i$ with positive floors; an independent success-calibration split gives
\begin{equation}
 d_n=\|S_i^{-1}(r_n^i-\mu_i)\|_\infty,\qquad
 q_i=\operatorname{Quantile}_{0.95}(\{d_n\}).
 \label{eq:quantile}
\end{equation}
Using ordered sample $\lceil0.95n_i\rceil$ and $D_i=q_iS_i$ (positive floor on $q_i$), define
\begin{equation}
 \begin{aligned}
 m_i(r)&=1-\|D_i^{-1}(r-\mu_i)\|_\infty,\\
 \mathcal R_i&=\{r:m_i(r)\geq0\},\quad
 \mathcal R_i^\delta=\{r:m_i(r)\geq\delta\}.
 \end{aligned}
 \label{eq:margin}
\end{equation}
Fitting and evaluation use disjoint scenes. P95 calibrates the reference region's empirical coverage. SCM verifies holding and support from observed evidence. At the checkpoint, valid base execution uses $\mathcal R_i$, and repair targets $\mathcal R_i^\delta$ with $0<\delta<1$.

\textbf{Candidate geometry and checkpoints.}
Gripper changes, low-speed locations, and kinematic directions identify checkpoints. FK of the unexecuted prefix gives a \emph{boundary residual trace} (BRT) for short-horizon geometric screening. A base prefix may continue if it can establish the relation before the checkpoint. Fresh observations verify the required condition before crossing.

\textbf{Runtime gain selection.}
For one predicted residual, RGC checks permission-respecting candidates from \eqref{eq:composition} over an ordered set $\mathcal A\subset(0,1]$:
\begin{equation}
 \mathcal A_t^{\mathrm{ok}}=\left\{\alpha\in\mathcal A:
 m_i(\widehat r_{\mathrm{end}}^{(\alpha)})\geq\delta,
 \;\mathcal K_t(A_t^{(\alpha)})=1\right\},
 \label{eq:feasible}
\end{equation}
Here $\mathcal K_t$ checks joint, step-size, and permission constraints. When a feasible gain exists, RGC selects $\min\mathcal A_t^{\mathrm{ok}}$, and the runtime executes its prefix. If the feasible set is empty, SCM retains affected holds and refreshes evidence or regenerates the correction. Gains and references are fixed before evaluation; checks repeat after new observations. Valid base execution continues under SCM permissions.

\begin{algorithm}[t]
\caption{Commitment Monitoring and Local Correction}
\label{alg:runtime}
\footnotesize
\noindent\textit{One call per monitoring cycle; history and unresolved commitments persist across calls.}\par\smallskip
\algrenewcommand{\algorithmicrequire}{\textbf{Input:}}
\algrenewcommand{\algorithmicindent}{1em}
\begin{algorithmic}[1]
\Require $A_{B,t}$, observation history, $E_i$, $(\mu_i,D_i)$, $\mathcal A$
\State Perceive and estimate semantic state (Secs.~\ref{sec:perception}--\ref{sec:runtime}).
\State Locate checkpoint; compute BRT (Sec.~\ref{sec:verification}).
\State Verify commitments; assign $\Gamma_t$ via \eqref{eq:permissions}.
\If{$\exists(k,d):\Gamma_{t,k,d}=\Correct$}
 \State $(c_t,\Delta A_t^c)\leftarrow\operatorname{BoundaryFlow}(z_t)$, \eqref{eq:head}.
 \State Form candidates and $\mathcal A_t^{\mathrm{ok}}$ via \eqref{eq:composition} and \eqref{eq:feasible}.
 \If{$\mathcal A_t^{\mathrm{ok}}\ne\varnothing$}
  \State $\alpha_t\leftarrow\min\mathcal A_t^{\mathrm{ok}}$; $\operatorname{ExecutePrefix}(A_t^{(\alpha_t)})$.
 \Else\ preserve valid motion; retain affected holds.
 \EndIf
\Else\ $\operatorname{ExecutePrefix}(A_t^{\mathrm{gate}})$ from \eqref{eq:composition}.
\EndIf
\State Refresh observations/history; re-verify before resuming held actions.
\State Advance only on event completion and successor readiness.
\end{algorithmic}
\end{algorithm}

\section{Experiments}
We evaluate the effects of commitment monitoring and local correction on task-level performance, their applicability across base policies, and their benefits over re-querying or inverse kinematics, and examine correction reuse across related interactions.

Table~\ref{tab:main} includes the shared benchmark results reported in CAPS~\cite{caps}.

\begin{table*}[!t]
\centering
\caption{RoboTwin 2.0 Results (\%)}
\label{tab:main}
\begingroup
\footnotesize\setlength{\tabcolsep}{2.6pt}\renewcommand{\arraystretch}{1.0}
\begin{tabular*}{\textwidth}{@{\extracolsep{\fill}}lcccccccccccc@{}}
\toprule
Method & \thd{Handover\\Block} & \thd{Handover\\Mic} & \thd{Hang\\Mug} & \thd{Bread\\Basket} & \thd{Bread\\Skillet} & \thd{Place A2B\\Left} & \thd{Place A2B\\Right} & \thd{Object\\Basket} & \thd{Object\\Stand} & \thd{Object\\Scale} & \thd{Object\\Cabinet} & Avg.$_{10}$\\
\midrule
RDT & 45 & \textbf{90} & 1 & 10 & 5 & 3 & 1 & 33 & 15 & 1 & 33 & 23.6\\
$\pi_{0.5}$ & 24 & 58 & 11 & 62 & 62 & 51 & 39 & 69 & 68 & 45 & 54 & 53.2\\
$\pi_{0.5}$+TACO~\cite{taco} & 36 & 63 & -- & 65 & 61 & 56 & 42 & 78 & 78 & 52 & 56 & 58.7\\
$\pi_{0.5}$+CAPS~\cite{caps} & 38 & 72 & -- & 63 & 69 & 54 & 47 & 76 & 81 & 57 & 58 & 61.5\\
$\pi_{0.5}$+CommitFlow & \textbf{66} & 60 & \textbf{42} & \textbf{75} & \textbf{78} & \textbf{70} & \textbf{69} & \textbf{88} & \textbf{97} & \textbf{85} & \textbf{71} & \textbf{75.9}\\
\bottomrule
\end{tabular*}
\endgroup
\par\smallskip
\begin{minipage}{\linewidth}
\footnotesize\raggedright
Task success rates over 100 trials per task with randomized scenes. Avg.$_{10}$ averages the 10 common tasks, excluding Hang Mug. Bold marks column best; ``--'' denotes an unreported value.
\end{minipage}
\end{table*}

\begin{figure*}[!t]
\centering
\setlength{\abovecaptionskip}{3pt}
\includegraphics[width=\textwidth]{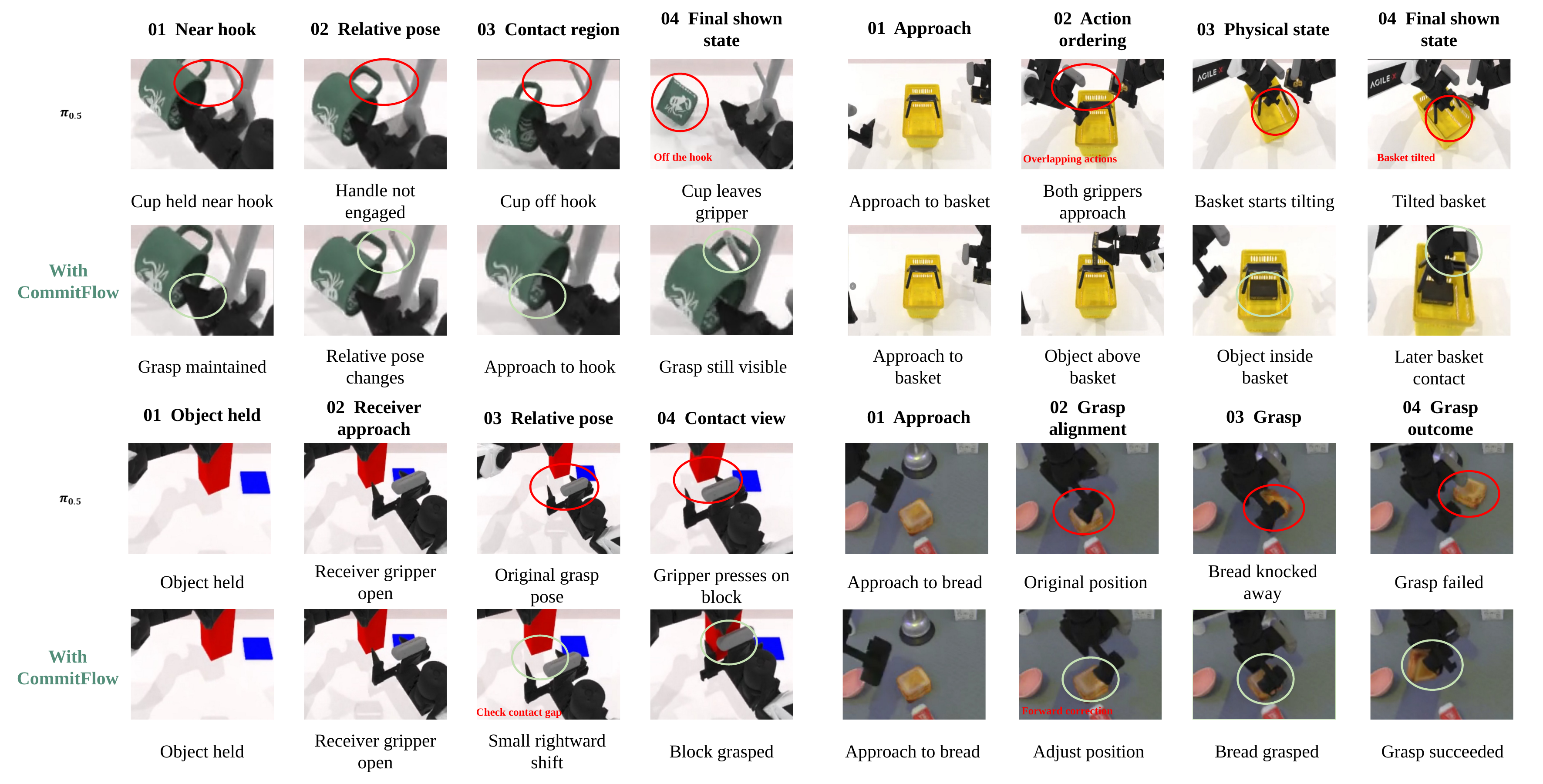}
\caption{\textbf{Qualitative execution sequences in simulation.} Four examples compare the base policy (upper rows) with CommitFlow (lower rows). Frames progress from left to right; marked regions highlight the relevant physical relations.}
\label{fig:execution-sequences}
\vspace{-4pt}
\end{figure*}

\subsection{Experimental Setup}
We evaluate 11 long-horizon interactive tasks from RoboTwin 2.0~\cite{robotwin}, covering handover, hanging, directional placement, and multi-target placement. Each task is evaluated over 100 trials with different random seeds. Table~\ref{tab:main} reports the mean success rate across 10 common tasks. The cross-policy evaluation in Table~\ref{tab:policies} averages all 11 tasks. The main comparisons include RDT~\cite{rdt}, $\pi_{0.5}$~\cite{pi05}, TACO, CAPS, and CommitFlow. We also integrate CommitFlow with $\pi_0$~\cite{pi0} and LingBot-VLA~\cite{lingbot} to assess its applicability across base policies. During CommitFlow training and evaluation, all base-policy parameters remain frozen, and the corresponding BoundaryFlow correction head is configured according to the base policy's action representation.

All policies share the relation-verification and action-permission logic; compatible correction heads adapt the output to each policy's action representation. BoundaryFlow uses a 10-step, six-joint window and a rank-12 basis.

\begin{figure}[!t]
\centering
\includegraphics[width=\columnwidth]{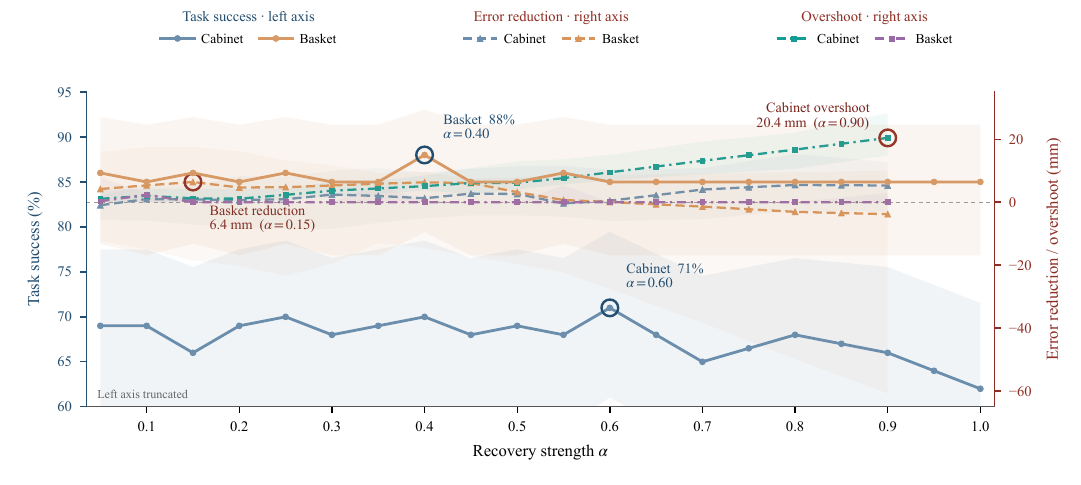}
\caption{\textbf{Recovery-strength sweep.} Cabinet and Basket: overall task success (left axis), recovery-error reduction and zero-crossing overshoot in mm (right axis). Circled annotations highlight selected measurements; the success axis is truncated.}
\label{fig:strength}
\end{figure}

\begin{figure*}[!t]
\centering
\includegraphics[width=\textwidth]{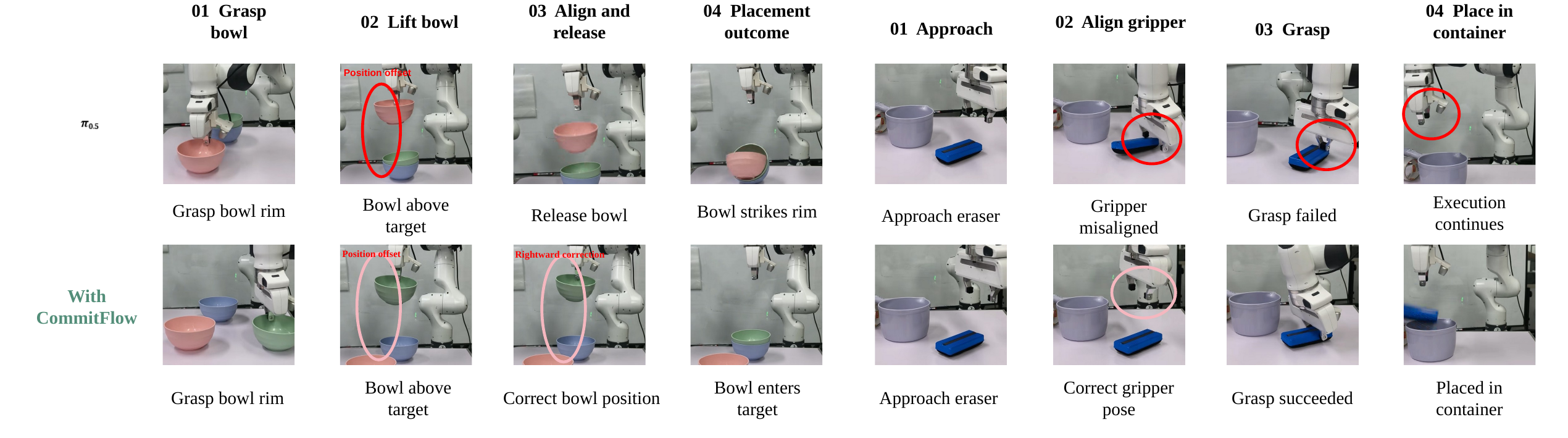}
\caption{\textbf{Qualitative execution sequences on the real robot.} Bowl stacking (left) and eraser placement (right), comparing the base policy (upper row) with CommitFlow (lower row). Frames progress from left to right.}
\label{fig:realworld-execution}
\end{figure*}

\begin{table}[!t]
\centering
\caption{Cross-Policy Results (\%)}
\label{tab:policies}
\begingroup
\footnotesize\setlength{\tabcolsep}{2pt}\renewcommand{\arraystretch}{1.0}
\begin{tabular*}{\columnwidth}{@{\extracolsep{\fill}}lccc@{}}
\toprule
Task & $\pi_0$~\cite{pi0} & $\pi_{0.5}$~\cite{pi05} & LingBot-VLA~\cite{lingbot}\\
\midrule
Handover Block & 26 $\rightarrow$ \textbf{45} & 24 $\rightarrow$ \textbf{66} & 85 $\rightarrow$ \textbf{90}\\
Handover Mic & 86 $\rightarrow$ \textbf{89} & 58 $\rightarrow$ \textbf{60} & 95 $\rightarrow$ \textbf{96}\\
Hang Mug & 3 $\rightarrow$ \textbf{25} & 11 $\rightarrow$ \textbf{42} & 49 $\rightarrow$ \textbf{62}\\
Bread Basket & 35 $\rightarrow$ \textbf{50} & 62 $\rightarrow$ \textbf{75} & 89 $\rightarrow$ \textbf{91}\\
Bread Skillet & 12 $\rightarrow$ \textbf{23} & 62 $\rightarrow$ \textbf{78} & 80 $\rightarrow$ \textbf{83}\\
Place A2B Left & 30 $\rightarrow$ \textbf{45} & 51 $\rightarrow$ \textbf{70} & 83 $\rightarrow$ 83\\
Place A2B Right & 28 $\rightarrow$ \textbf{38} & 39 $\rightarrow$ \textbf{69} & 82 $\rightarrow$ \textbf{84}\\
Object Basket & 50 $\rightarrow$ \textbf{73} & 69 $\rightarrow$ \textbf{88} & 73 $\rightarrow$ \textbf{94}\\
Object Stand & 60 $\rightarrow$ \textbf{80} & 68 $\rightarrow$ \textbf{97} & 95 $\rightarrow$ \textbf{96}\\
Object Scale & 53 $\rightarrow$ \textbf{73} & 45 $\rightarrow$ \textbf{85} & 96 $\rightarrow$ 93\\
Object Cabinet & 18 $\rightarrow$ \textbf{41} & 54 $\rightarrow$ \textbf{71} & 68 $\rightarrow$ \textbf{81}\\
\midrule
Avg.$_{11}$ & 36.5 $\rightarrow$ \textbf{52.9} & 49.4 $\rightarrow$ \textbf{72.8} & 81.4 $\rightarrow$ \textbf{86.6}\\
\bottomrule
\end{tabular*}
\endgroup
\par\smallskip
\begin{minipage}{\linewidth}
\footnotesize\raggedright
Success rates: Base $\rightarrow$ Base+CommitFlow. Bold marks improvement. Avg.$_{11}$ averages all 11 tasks. Each policy uses a compatible correction head.
\end{minipage}
\end{table}

\subsection{Task Performance}
Table~\ref{tab:main} reports a ten-task mean success rate of 75.9\% for CommitFlow, compared with 53.2\%, 58.7\%, and 61.5\% for $\pi_{0.5}$, TACO, and CAPS. Improvements span directional placement, object placement, and handover; for example, Handover Block increases from 24\% to 66\%. These results support verifying physical relations and applying local correction before execution errors propagate to later stages.

Table~\ref{tab:policies} reports results across all 11 tasks: Hang Mug rises from 3\% to 25\% with $\pi_0$, while Object Basket and Object Cabinet increase from 73\% to 94\% and 68\% to 81\% with LingBot-VLA. Gains are not universal: LingBot-VLA remains at 83\% on Place A2B Left and decreases from 96\% to 93\% on Object Scale. The shared execution mechanism thus improves performance without guaranteeing gains on every policy--task pair.

Figure~\ref{fig:execution-sequences} compares the base policy and CommitFlow in four simulated examples: mug--hook alignment, placement--grasp ordering, block handover, and bread grasping. Successive observations are arranged from left to right.

\begin{table}[!t]
\centering
\caption{Local Correction Ablation (\%)}
\label{tab:fresh}
\begingroup
\footnotesize\setlength{\tabcolsep}{2.5pt}\renewcommand{\arraystretch}{1.0}
\begin{tabular*}{\columnwidth}{@{\extracolsep{\fill}}lccc@{}}
\toprule
Task & Fresh Base & Fresh + IK & Fresh + Ours\\
\midrule
Object Cabinet & 61 & 57 & \textbf{71}\\
Object Basket & 60 & 68 & \textbf{88}\\
Place A2B Left & 55 & 57 & \textbf{70}\\
Place A2B Right & 56 & 60 & \textbf{69}\\
Object Scale & 78 & 78 & \textbf{85}\\
Object Stand & 91 & 94 & \textbf{97}\\
\midrule
Avg. & 66.8 & 69.0 & \textbf{80.0}\\
\bottomrule
\end{tabular*}
\endgroup
\par\smallskip
\begin{minipage}{\linewidth}
\footnotesize\raggedright
Task success rates. All variants re-query Base; Avg. is the six-task macro-average.
\end{minipage}
\end{table}

\subsection{Ablation Experiments}
We ablate the local-correction method and correction strength. With all variants re-querying the base policy, Fresh + Ours achieves 80.0\% mean success across six tasks, versus 66.8\% for Fresh Base and 69.0\% for Fresh + IK (Table~\ref{tab:fresh}). IK reduces Object Cabinet success from 61\% to 57\%, suggesting that geometric alignment alone may not restore compatibility with subsequent base-policy actions.

Figure~\ref{fig:strength} reports 71\% success on Cabinet at $\alpha=0.60$ and 88\% on Basket at $\alpha=0.40$. Larger correction strengths increase Cabinet overshoot, reaching 20.4\,mm at $\alpha=0.90$. This motivates balancing relation recovery against disruption of effective base motion. RGC scales BoundaryFlow's predicted correction direction according to the current state, so its corrective capacity remains bounded by the local residual representation. Such corrections may be insufficient for complex contact dynamics, obstacle avoidance, regrasping, or global trajectory planning.

Among recovery-triggered samples involving regrasping after alignment correction, local recovery and final task success rates are 45.8\% (109/238) and 18.1\% (43/238) for $\pi_0$, and 66.7\% (100/150) and 46.7\% (70/150) for $\pi_{0.5}$. Thus, successful local grasp recovery does not guarantee success in subsequent task stages. CommitFlow repairs local deviations while preserving valid progress, but cannot provide task skills absent from the base policy.

\subsection{Local-Correction Reuse Across Objects}
We evaluate whether learned local corrections can be reused across tasks with the same interaction structure but different objects. For handover, a BoundaryFlow head trained only on Handover Block correction samples is applied to Handover Mic without target-task correction demonstrations. The correction head remains unchanged, while the object--receiver relation reference is rebound to the new interaction. Similarly, for Can $\rightarrow$ Basket, we reuse the correction head trained on Object Basket and bind it to the corresponding can--basket relation.

Table~\ref{tab:reuse-realworld} summarizes both correction-reuse evaluations in simulation. Success increases from 58\% to 65\% on Handover Mic and from 40\% to 58\% on Can $\rightarrow$ Basket. These results indicate that relation-conditioned local corrections can remain effective after object substitution, supporting reuse of an existing correction head across related interactions rather than requiring a separately trained correction module for every object configuration.

\begin{table}[!ht]
\centering
\caption{Correction Reuse and Real-World Results (\%)}
\label{tab:reuse-realworld}
\begingroup
\footnotesize\setlength{\tabcolsep}{4pt}\renewcommand{\arraystretch}{1.0}
\begin{tabularx}{\columnwidth}{@{}Xcc@{}}
\toprule
Task & Base & + CommitFlow\\
\midrule
\multicolumn{3}{@{}l}{\textbf{Correction reuse --- simulation}}\\
Handover Mic & 58 & \textbf{65}\\
Can $\rightarrow$ Basket & 40 & \textbf{58}\\
\midrule
\multicolumn{3}{@{}l}{\textbf{Real-world execution}}\\
Eraser $\rightarrow$ Container & 50 & \textbf{63}\\
Stack Three Bowls & 40 & \textbf{50}\\
\bottomrule
\end{tabularx}
\endgroup
\par\smallskip
\begin{minipage}{\linewidth}
\footnotesize\raggedright
Task success rates. Base denotes $\pi_{0.5}$.
\end{minipage}
\end{table}

\subsection{Real-World Experiments}
We further evaluate CommitFlow on two real-world manipulation tasks, with 100 trials per task. As shown in Table~\ref{tab:reuse-realworld}, CommitFlow improves the success rate of $\pi_{0.5}$ from 50\% to 63\% on Place Blackboard Eraser in Container (Eraser $\rightarrow$ Container) and from 40\% to 50\% on Stack Three Bowls.

Figure~\ref{fig:realworld-execution} further illustrates how these improvements arise during execution. In bowl stacking, the base policy releases while the bowl remains offset from the target, causing contact with the rim, whereas CommitFlow corrects the bowl position before release. In eraser placement, gripper misalignment causes the base policy to miss the grasp, while CommitFlow adjusts the gripper pose before closure and subsequently completes the placement. These examples illustrate pre-transition intervention: local geometric deviations are corrected before critical dependent actions such as grasp closure or release are allowed to proceed.

\section{Conclusion}
We presented CommitFlow, which verifies semantic commitments and combines local correction with gain calibration before critical stage transitions. Across simulation and real-world tasks, the framework improves execution with frozen base policies, while correction-reuse experiments support transfer across related interactions. Its central principle is to repair unmet physical conditions while preserving valid progress and to re-verify those conditions using fresh observations before dependent actions resume.

Runtime verification remains dependent on RGB-D perception and temporal state estimation. Occlusion, invalid depth, object-state estimation errors, and temporal asynchrony can compromise the estimated relations and physical conditions. Task-dependent event and precondition definitions also limit extension to open-ended tasks and unfamiliar interaction relations. Future work will explore adaptive task adapters that infer event structures, manipulation roles, and physical preconditions from observations and language instructions. Execution feedback will guide adapter refinement and condition verification under perceptual uncertainty. We will also investigate online calibration of stage reference regions and verification thresholds using verified execution outcomes. Conservative updates will aim to accommodate changing objects and sensing conditions while preserving the closed loop of verification, correction, and re-verification.

\FloatBarrier
\balance
\bibliographystyle{IEEEtran}
\bibliography{references}
\end{document}